\documentclass[lettersize,journal]{IEEEtran}
\pdfoutput=1 
\usepackage{float}
\pdfoutput=1
 
\usepackage{cite}
\usepackage[pdftex]{graphicx}
\usepackage[section]{placeins}
\usepackage{tabularx}
\usepackage{siunitx}
\usepackage{amsmath, amssymb,mathrsfs, dsfont} 
\usepackage{amsmath,amssymb,bm,bbm}
\usepackage{graphics,graphicx}

\usepackage{multirow}
\usepackage{hyperref}
\usepackage{soul}

\usepackage{xfrac}  

\usepackage{amsmath}
\usepackage{subfiles} 
\usepackage{booktabs}
\usepackage{amsfonts}
\usepackage{booktabs}
\usepackage{siunitx}
\usepackage[dvipsnames, table]{xcolor}
\usepackage[deletedmarkup=sout,authormarkup=superscript]{changes}
\definechangesauthor[name={Mahdi}, color=NavyBlue]{MN}

\graphicspath{{\subfix{../figures/}}}

\usepackage{lineno}

\usepackage{booktabs} 
\definecolor{fadedgray}{rgb}{0.8, 0.8, 0.8}  

\usepackage{amsthm} 

\usepackage{tikz}
\usetikzlibrary{arrows.meta,calc}

\usepackage{algorithmicx}
\usepackage{algcompatible}
\usepackage{subcaption}  
\usepackage{algorithm} 
\usepackage{algpseudocode}

\begin{document}
\title{Residual Reinforcement Learning for Visual Servo Control with Zero-Shot Sim-to-Real Transfer}
\author{Mahdi Nobar\IEEEauthorrefmark{1},
        Alisa Rupenyan\IEEEauthorrefmark{2},
        John Lygeros \IEEEauthorrefmark{3},  
        and J\"urg Keller\IEEEauthorrefmark{4}
\thanks{The Swiss National Science Foundation supported this work through NCCR Automation, grant number 180545. (Corresponding author: Mahdi Nobar)}
\thanks{\IEEEauthorrefmark{1}Mahdi Nobar and \IEEEauthorrefmark{3}John Lygeros are with the Automatic Control Laboratory, ETH Z\"urich, Zurich, Switzerland (email: mnobar@ethz.ch; jlygeros@ethz.ch).}
\thanks{\IEEEauthorrefmark{2}Alisa Rupenyan is with the ZHAW Center for AI, Zurich University of Applied Sciences, Winterthur, Switzerland (email: alisa.rupenyan@zhaw.ch).}
\thanks{\IEEEauthorrefmark{3}J\"urg Keller is with the Automation Institute, FHNW, Windisch, Switzerland (email: juerg.keller1@fhnw.ch).}}

\maketitle
\pagestyle{plain}
\begin{abstract}
We address tracking a moving object on a conveyor with a Franka Emika Panda robot arm, despite an imprecise digital twin, measurement errors, and noise.
Our novel approach augments a loop-shaped inverse–Jacobian PI controller with a residual policy learned via Soft Actor–Critic in the digital twin.
A delay-compensated Kalman filter fuses out-of-sequence camera measurements to provide latency-aware moving target position estimates.
We compare the achieved performance to a performance bound for a class of model-based controllers derived from a frequency-domain sensitivity analysis, quantifying the minimum achievable error imposed by uncertain kinematic model mismatch.
Simulation and hardware experiments indicate that our residual policy drives tracking error below the bound. 
Across conveyor speeds ($27.9$--$41.8\ \mathrm{mm/s}$), our method reduces mean tracking errors from $1.1$--$1.4\ \mathrm{mm}$ (baseline inverse-Jacobian PI) to $0.3$--$0.7\ \mathrm{mm}$. 
These results show that learning residual corrections in a limited-fidelity simulator, combined with delay compensation, can substantially improve model-based control under modeling and measurement uncertainties while maintaining reliable sim-to-real transfer.
\end{abstract}
\begin{IEEEkeywords}
Reinforcement Learning, Sim-to-Real Transfer, Kalman Filter, Soft Actor–Critic, Robot Tracking Control
\end{IEEEkeywords}
\IEEEpeerreviewmaketitle

\section{Introduction}\label{sec:I_introduction}
Autonomous manipulation on moving conveyor belts is a pivotal challenge in industrial automation because unmodeled belt and robot dynamics, sensor delays, and kinematic uncertainties hinder precise tracking.
Classical model-based controllers guarantee stability yet degrade under model bias or localization noise \cite{rana2023bayesian}.
Pure reinforcement learning (RL) approaches are adaptive but data-intensive and impractical to train exclusively on hardware \cite{10679148}.
Hybrid methods alleviate these limitations by combining model-based and data-driven control.
Residual learning enables an RL agent to augment the nominal controller, leveraging model knowledge for stability and sample efficiency while locally compensating for modeling and sensing errors \cite{10689563}.
Learning-based model predictive control has likewise achieved high-precision trajectory tracking on compliant robotic arms by integrating a Gaussian process–based model correction and an online disturbance estimator \cite{8768048}.
Similarly, data-driven trajectory optimization has been shown to improve motion accuracy in precision systems without modifying the underlying low-level controllers \cite{BALULA2024105834}.
Researchers have also addressed the simulation–to-reality gap by, for example, pairing simulated pretraining with domain randomization across varied parameters \cite{zhang2024catch} or constraining policy updates near nominal trajectories to improve transferability and convergence \cite{ma2025online}.
Alternative iterative approaches address model mismatch through repetitive control and run-to-run optimization \cite{10666830}, while data-efficient optimization with digital twins accelerates controller tuning by substituting costly real experiments with simulated trials \cite{nobar2025guidedmultifidelitybayesianoptimization}.
Learning-based iterative controllers achieve high-precision tracking through compensation across repeated executions \cite{balta2021learning,wu2025iterative}, and adaptive Bayesian optimization uses Gaussian process models to improve performance over multiple runs \cite{konig2025adaptive}.
In dynamic grasping and conveyor picking, deterministic planners predict object motion for interception \cite{carlier2025performance}, adaptive controllers reconfigure trajectories in real-time \cite{10711690}, and multi-agent RL with replay optimization enables interception on moving platforms \cite{HUANG2024102644}.

We propose a control framework that augments a nominal inverse-Jacobian PI controller with a data-driven residual Soft Actor-Critic (SAC) policy to address autonomous manipulation on moving conveyor belts.
The proposed framework contributes three elements: (i) a principled architecture that integrates model-based control and reinforcement learning for real-time refinement of the control signal through zero-shot sim-to-real transfer, (ii) a varying delay-compensated Kalman filter for vision-based state estimation under sensing latency, and (iii) a power-spectral tracking error lower bound derived from the closed-loop sensitivity.
The SAC agent is trained entirely in simulation without any real-system data and compensates for modeling and sensing uncertainties, including kinematic bias, unmodeled dynamics, belt flexibility, measurement latency, and noise, by refining the commands generated by the nominal controller. 
We validate the proposed controller on a real robot using a vision-based robot cell with a dead–time–compensated Kalman filter and digital twin simulations, achieving sub-millimeter tracking accuracy that surpasses the performance limit on a class of model-based baseline controllers across varying conveyor speeds.

Section \ref{sec:II_problem_statement} defines the problem and control objective; Section \ref{sec:III_Residual_Reinforcement_Learning} presents the proposed controller algorithm; Section \ref{sec:IV_lower_bound} derives a lower bound for the tracking error of the class of controllers; and Section \ref{sec:V_performance_evaluation} discusses simulation results and experimental validations.


\section{Problem Statement}\label{sec:II_problem_statement}
We aim to track a moving object on a conveyor belt using a fixed gripper joint on a 6-degree-of-freedom serial robot manipulator.
Let $\dot{\mathbf{q}}{\text{c}}(k)\in\mathbb{R}^6$ denote the commanded joint velocity at discrete time index $k \in \mathcal{W}$.
The \textit{real system} provides noisy measurements of the joint positions $\mathbf{q}(k)\in\mathbb{R}^6$, joint velocity $\dot{\mathbf{q}}(k)\in\mathbb{R}^6$, and end-effector position $\mathbf{p}(k)\in\mathbb{R}^3$.
We assume a \textit{digital twin} of the physical plant exists, based on the identified model in \cite{Panda_identify_8772145}, with limited accuracy.
Given the commanded joint velocity, the digital twin predicts joint positions, joint velocities, joint torques, and end-effector position, all subject to errors due to its limited accuracy.
We assume the robot's kinematic model has a bias that depends on the robot configuration $\mathbf{q}(k)$ at time index $k$.
Let $J\bigl(\mathbf{q}(k)\bigr)\in\mathbb{R}^{3\times6}$ denote the true translational Jacobian and $J^{\dagger}\bigl(\mathbf{q}(k)\bigr)\in\mathbb{R}^{6\times3}$ its pseudo-inverse.
The \textit{biased} Jacobian pseudo-inverse is defined as
\begin{equation}
\hat{J}^{\dagger}\bigl(\mathbf{q}(k)\bigr)
:= J^{\dagger}\bigl(\mathbf{q}(k)\bigr)
+ \Delta_{a}\bigl(\mathbf{q}(k)\bigr),
\label{eq:hpiJ}
\end{equation}
where $\Delta_{a}\bigl(\mathbf{q}(k)\bigr)\in\mathbb{R}^{6\times3}$ is the additive uncertainty capturing unstructured kinematic model bias at joint position $\mathbf{q}(k)$.

\begin{figure}[!t]
\centering
\includegraphics[width=.8\columnwidth, trim=3.3cm 0.1cm 3.3cm 0.1cm, clip]{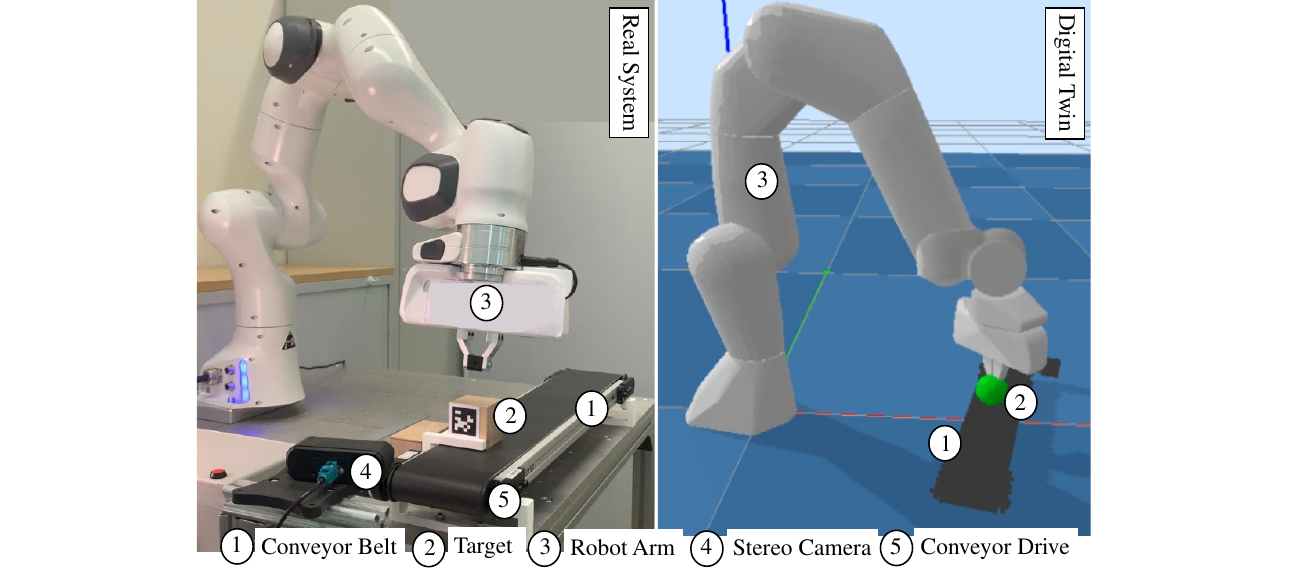}
\caption{Components of the closed-loop robot conveyor tracking system and its digital twin}
\label{fig:FEP_setup}
\end{figure}

A vision-based motion planner fuses delayed and noisy stereo-camera target position measurements $\mathbf{p}^\in \mathbb{R}^3$ with noisy conveyor drive belt speed feedback $\dot{\mathbf{p}}^\in \mathbb{R}^3$ to estimate the target position $\tilde{\mathbf{p}}^{*}(k)\in\mathbb{R}^3$.
Fig. \ref{fig:FEP_setup} shows the hardware setup and its digital twin.

The tracking error in the Cartesian coordinate system is defined as
\begin{equation}\label{eq:e}
\mathbf{e}(k):=\tilde{\mathbf{p}}^{*}(k)-\mathbf{p}(k),
\end{equation}
where $\mathbf{e}(k)=[e_x(k),e_y(k),e_z(k)]^{\top}\in \mathbb{R}^3$.
Given the limited accuracy of the digital twin with biased kinematic and dynamic models, the control objective is to compute $\dot{\mathbf{q}}{\mathrm{c}}(k)\in \mathbb{R}^6$ such that the end-effector accurately tracks the moving target on the real robot system, based on uncertain target position estimates, by minimizing the tracking error:
\begin{equation}
\label{eq:objective_formulation}
\min_{\dot{\mathbf{q}}{\mathrm{c}}(k)}\; \|\mathbf{e}(k)\|,
\end{equation}
where $\|\cdot\|$ is the Euclidean norm.

\section{Residual Reinforcement Learning}\label{sec:III_Residual_Reinforcement_Learning}
\begin{figure}[!h]
\centering
\includegraphics[width=.9\columnwidth, trim=6.2cm 0.1cm 6.2cm 0cm, clip]{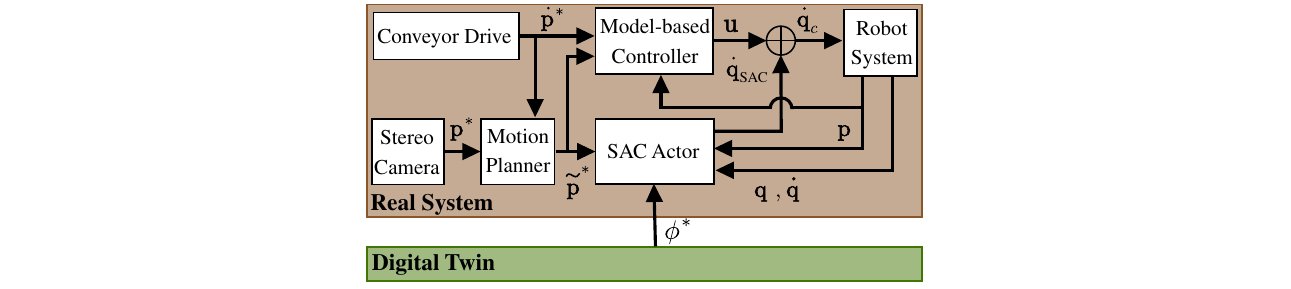}
\caption{Residual learning control framework combining SAC and model-based controller for zero-shot transfer on real robot\label{fig:method_overview}}
\end{figure}

Our approach trains a reinforcement learning controller using the SAC algorithm within the robot's digital twin.
We deploy the trained SAC policy on the real system in a \textit{zero-shot} manner, that is, without any further fine-tuning on hardware.
The model-based controller, designed using the biased kinematic model, provides the nominal joint–velocity command $\mathbf{u}(k)\in \mathbb{R}^6$.
At the same time, the SAC policy $\pi_{\phi^*}$ trained in digital twin with the optimum parameters $\phi^*$ generates an additive correction $\dot{\mathbf{q}}_{\mathrm{SAC}}(k)\in \mathbb{R}^6$. 
The resulting control input applied to the real robot is 
\begin{equation}\label{eq:dq_c}
\dot{\mathbf{q}}_{\mathrm{c}}(k) := \mathbf{u}(k) + \dot{\mathbf{q}}_{\mathrm{SAC}}(k).
\end{equation}
We refer to the residual SAC--augmented inverse--Jacobian PI controller as the \emph{residual SAC--PI controller}.
Fig. \ref{fig:method_overview} shows an overview of the framework. 
The remainder of this section describes the closed-loop system and the nominal model-based controller, followed by the vision-based motion planner and training of the residual SAC controller.

\subsection{Closed-Loop System Description}\label{subsec:closed_loop_system}
\begin{figure}[!h]
\centering
\includegraphics[width=.99\columnwidth, trim=4.2cm 0.1cm 4.2cm 0cm, clip]{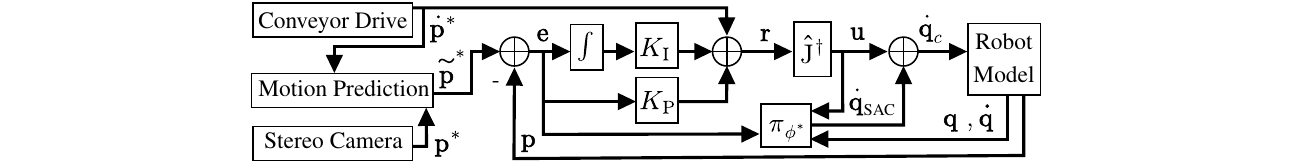}
\caption{Closed-loop system with an inverse--Jacobian PI controller.}
\label{fig:closed_loop}
\end{figure}

We consider a model-based inverse--Jacobian PI controller that, at each control step, computes the joint-velocity command $\mathbf{u}(k)\in\mathbb{R}^{6}$ according to
\begin{align}
\mathbf{r}(k) &:= \dot{\mathbf{p}}^{*}(k)
                 + K_{\mathrm{P}}\mathbf{e}(k)
                 + K_{\mathrm{I}}\!\sum_{i=0}^{k}\mathbf{e}(i), \label{eq:r}\\
\mathbf{u}(k) &:= \hat{J}^{\dagger}\!\bigl(\mathbf{q}(k)\bigr)\,\mathbf{r}(k), \label{eq:u}
\end{align}
where $K_{\mathrm{P}},K_{\mathrm{I}}\in\mathbb{R}^{3\times 3}$ are diagonal gain matrices.
Fig.~\ref{fig:closed_loop} illustrates the resulting closed-loop system.

Let $P$ denote the overall plant, including the robot arm, low-level controller, actuators, and encoders.
The nominal loop transfer function, in which there is no bias in the kinematic model, is defined as
\begin{equation}\label{eq:L0}
L_{0}(e^{j\omega}) := P(e^{j\omega}) J^{\dagger}(\mathbf{q}(k)) C(e^{j\omega}),
\end{equation}
where $C(e^{j\omega}) := K_{\mathrm{P}} + K_{\mathrm{I}} G(e^{j\omega})$ and $G(e^{j\omega})$ comprises three discrete-time integrators.

\subsection{Model-based Controller Gains Optimization}
\label{subsec:gain_optim}

We derive a model for each robot joint to design the inverse--Jacobian PI controller gains $\bigl(K_{\text{P}}, K_{\text{I}}\bigr)$ in \eqref{eq:r} using classical loop shaping.

\subsubsection{Simplified Plant Model}
We identify a simplified joint model, denoted by $G_q(e^{j\omega}):= \operatorname{diag}\!\bigl(g_i(e^{j\omega})\bigr)_{i=1}^{6} \in \mathbb{C}^{6\times 6}$ for each joint, to capture actuation dynamics through a reduced-order representation. 
The control input is $\mathbf{u}(k) := \bigl(u_i(k)\bigr)_{i=1}^{6}$, where each joint’s discrete-time response from $u_i(k)$ to $\dot{q}_i(k)$ is modeled as a first-order transfer function $g_i(e^{j\omega}) := \frac{b_{i}}{e^{j\omega} - a_i}$ such that
\begin{align*}
\dot{q}_i(k)
&:= a_i\,\dot{q}_i(k-1) + b_{i}\,u_i(k-1),\\
a_i &:= e^{-\delta t / \tau_i}, \qquad
b_{i} := 1 - a_i,
\end{align*}
where $\delta t>0$ and $\tau_i>0$ are the sampling period and the time constant of the $i$th joint, respectively.
In our digital twin, we apply a pseudo-random binary sequence of joint speed commands and identify the time constants $\boldsymbol{\tau} = [\,0.001,\; 0.006,\; 0.011,\; 0.097,\; 0.002,\; 0.008\,]^{\top}\!\ \mathrm{s}$ using least-squares fitting.

\subsubsection{Controller Gain Design}\label{subsubsec:controller_gains_design}
Assuming operation near a nominal configuration $\mathbf{q}(0)$, the geometric Jacobian is locally denoted as $J\bigl(\mathbf{q}(0)\bigr)$.
The controller gains are optimized using the simplified discrete-time plant model 
\begin{equation}\label{eq:P_simple_identified}
P(e^{j\omega}) = G(e^{j\omega}) \,J\bigl(\mathbf{q}(0)\bigl)\,G_q(e^{j\omega}) .
\end{equation}
The proportional and integral gains $K_{\mathrm{P}}$ and $K_{\mathrm{I}}$ are tuned using classical loop-shaping principles \cite{skogestad2005multivariable}.
For each Cartesian axis, a scalar equivalent loop $L_i(e^{j\omega})$ approximates the multivariable dynamics, and the gains are optimized to achieve the desired crossover frequency and phase margin.
The target crossover frequency is set to $(4\times \tfrac{1}{6}\sum_{i=1}^{6}\tau_i)^{-1}$, which balances phase margin and disturbance rejection for first-order actuator dynamics. 
According to the symmetric optimum criterion \cite{HAGGLUND1991727}, the proportional and integral gains $K_{\mathrm{P}}$ and $K_{\mathrm{I}}$ are jointly determined to achieve this target bandwidth, yielding zero steady-state error, a phase margin of approximately $60^{\circ}$, and a gain margin above $6\ \mathrm{dB}$. 
The resulting diagonal gains, $K_{\mathrm{P}}=\mathrm{diag}(5.6,6.8,5.6)$ and $K_{\mathrm{I}}=\mathrm{diag}(0.38,8.25,4.64)$, produce a closed-loop system with robust stability, smooth transients, and bandwidth consistent with the identified joint dynamics.

\subsection{Dead-Time Compensated Kalman Filter}
\label{subsec:kalman_filter}

We assume the camera measurements have a known timestamp $t_{\mathrm{meas}}$.
Due to variable image-processing and communication latencies, each measurement of the stereo camera arrives after a random delay $\Delta t:= t_{\mathrm{now}} - t_{\mathrm{meas}}$, where $t_{\mathrm{now}}$ denotes the arrival time.
When we are at time index $k$, the stereo camera provides delayed measurements of the target position $\mathbf{p}^{*}(k-d)\in\mathbb{R}^3$, at time index $k-d$ such that
\begin{equation}\label{eq:d}
d := \lfloor\frac{\Delta t}{\delta t}\rfloor,
\end{equation}
and $0 < d < k$.

\begin{figure}[!h]
\centering
\includegraphics[width=.99\columnwidth, trim=4.4cm 0.1cm 4.4cm 0cm, clip]{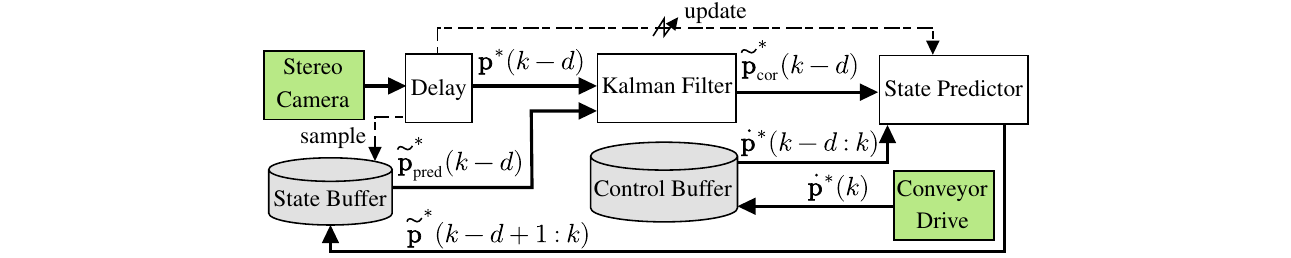}
\caption{Delay-compensated Kalman filter: each delayed camera measurement $\mathbf{p}^{*}(k-d)$ is inserted at its true acquisition index $k-d$, corrected once, and then propagated forward to the current control time $k$ to yield a prediction $\tilde{\mathbf{p}}^{*}(k)$.}
\label{fig:KF_delay_comp_1}
\end{figure}

To compensate for this delay, following \cite{bar2004estimation, maybeck1979stochastic}, we propose a dead-time-compensated Kalman filter (KF) with measurement insertion and forward propagation, as shown in Fig. \ref{fig:KF_delay_comp_1}.

When a delayed measurement $\mathbf{p}^{*}(k-d)$ arrives, the KF performs a single state correction at the corresponding buffered index $k-d$:
\begin{align*}
\boldsymbol{\nu}(k-d) &:= \mathbf{p}^{*}(k-d) - \tilde{\mathbf{p}}^{*}_{\text{pred}}(k-d),\\
K(k-d) 
&:= \Sigma_{\text{pred}}(k-d)\,\mathbb{I}_3^{\top}\\
&\quad\times
    \Bigl[
        \mathbb{I}_3\,\Sigma_{\text{pred}}(k-d)\,\mathbb{I}_3^{\!\top}
        + R(k-d)
    \Bigr]^{-1},\\
\tilde{\mathbf{p}}^{*}_{\text{cor}}(k-d) &:= \tilde{\mathbf{p}}^{*}_{\text{pred}}(k-d) + K(k-d)\,\boldsymbol{\nu}(k-d),\\
\Sigma_{\text{cor}}(k-d) &:= \bigl[I - K(k-d)\,\mathbb{I}_3\bigr]\,\Sigma_{\text{pred}}(k-d),
\end{align*}
where $(\cdot)_{\text{pred}}$ and $(\cdot)_{\text{cor}}$ denote the predicted and corrected quantities, $\boldsymbol{\nu}(k-d)$ is the innovation, $K(k-d)$ is the Kalman gain, $R(k-d)$ is the measurement-noise covariance, and $\mathbb{I}_3 \in \mathbb{R}^{3\times 3}$ is the measurement matrix, since the camera directly observes the target position.
After this correction, we perform forward propagation from the corrected state up to the current index $k$ using a linear model driven by the conveyor speed $\dot{\mathbf{p}}^*$ from the control buffer:
\begin{equation}
\tilde{\mathbf{p}}^{*}(i{+}1) = \tilde{\mathbf{p}}^{*}(i){+}\begin{bmatrix}0\\ \delta t\\ 0\end{bmatrix}\dot{\mathbf{p}}^{*}(i),\; i\in\{k-d,\dots,k-1\},
\end{equation}
where $\tilde{\mathbf{p}}^{*}(k-d)=\tilde{\mathbf{p}}^{*}_{\text{cor}}(k-d)$.
\subsection{Residual SAC Controller}
\label{subsec:residual_rl}

SAC is an off-policy actor-critic algorithm within the maximum-entropy framework, which has been shown to achieve state-of-the-art performance on continuous control tasks \cite{SAC2019}.
We model the closed-loop control problem as a Markov decision process, where at each discrete time index $k$, the agent observes a state $s(k)$, selects an action $a(k)$ according to its stochastic policy $\pi_{\phi}(a\,|\,s)$ parameterized by the actor network $\phi$, and receives a scalar reward $r(k)$.
The state, action, and reward are defined:
\begin{align}
s(k) &:= \bigl[\,\mathbf{e}(k),\, \mathbf{q}(k),\, \dot{\mathbf{q}}(k),\, \mathbf{u}(k),\, \dot{\mathbf{q}}_{\mathrm{SAC}}(k{-}1)\,\bigr] \in \mathbb{R}^{27},\\
a(k) &:= \dot{\mathbf{q}}_{\mathrm{SAC}}(k) \in \mathbb{R}^{6},\\
r(k) &:= \sum_{i\in\{x,y,z\}} \frac{L}{e^{l\,|e_i(k)|}+e^{-l\,|e_i(k)|}} \in \mathbb{R}.
\label{eq:reward}
\end{align}
The reward function encourages accurate tracking by assigning higher values for smaller position errors through the smooth logistic shaping parameters $L=2$ (scale) and $l=600$ (sensitivity).
The method uses two critics to mitigate value overestimation by taking the minimum of their estimates in the Bellman target and policy objective~\cite{SpinningUp2018}.
The critics and the squashed-Gaussian actor, with learned state-dependent mean and standard deviation, use the standard SAC parameterizations and losses~\cite{SAC2019};
Table~\ref{tab:sac_parameters} specifies their training configuration.
During training, the sampled residual joint commands from $\pi_{\phi}$ are clipped element-wise to $\pm \rho\,\dot{\mathbf{q}}_{\max}$ with $\rho := 0.1$ to limit residual authority during exploration.
SAC maximizes the discounted entropy-regularized objective
\begingroup
\begin{equation}
\max_{\phi}\;\mathbb{E}_{\pi_{\phi}}\!\left[
\sum_{k=0}^{\infty}\gamma^k\!\left(r(k)
-\alpha\log\pi_{\phi}\!\big(a(k)\,|\,s(k)\big)\right)\right],
\end{equation}
\endgroup
where $\gamma$ is the discount factor and $\alpha>0$ balances reward and entropy.
A higher $\alpha$ promotes exploration, while a lower $\alpha$ enforces more deterministic behavior;
\(\alpha\) is adapted toward the heuristic target entropy $-\dim(a)=-6$, automatically regulating exploration~\cite{SAC2019}.
The optimal parameters $\phi^{*}$ define the deterministic policy to generate the residual joint-speed commands $\dot{\mathbf{q}}_{\mathrm{SAC}}(k) = \pi_{\phi^{*}}(s(k))$.

To improve the sim-to-real gap, we implement domain randomization in our simulation environment.
Here, $\mathcal{N}(\mu,\sigma)$ denotes a Gaussian with mean $\mu$ and standard deviation $\sigma$.
During training, we randomize the kinematic model parameters at the start of each episode.
Specifically, each link's position and orientation are perturbed by random offsets drawn from $\mathcal{N}(0, 0.5\ \mathrm{mm})$ and $\mathcal{N}(0, 3^\circ)$, respectively, applied once at episode initialization, reflecting possible mismatch between real hardware and the kinematic model.
Additive zero-mean Gaussian noises are applied to both reference target positions and velocities.
We simulate the \textit{startup phase} in which the robot starts moving to the vicinity of the initial end-effector position from zero velocity up to near conveyor belt velocity with a smooth target speed profile.
At the end of the startup phase, we initialize the \textit{tracking phase} with joint positions and speeds $\mathbf{q}(0)$ and $\dot{\mathbf{q}}(0)$, respectively.
The initial speeds are perturbed by $\mathcal{N}(0,0.001)\ \mathrm{rad/s}$ once per episode.
Each episode consists of data from an entire tracking phase.
The initial joint angles are randomized as $\mathbf{q}(0) \leftarrow \mathbf{q}(0) + \mathcal{N}(0, 0.01)\ \mathrm{rad}$ and the measured joint velocities are perturbed at each time step $k$ as $\dot{\mathbf{q}}(k) \leftarrow \dot{\mathbf{q}}(k) + \mathcal{N}(0, 0.04)$ [rad/s].
Target is on a conveyor belt along the y-axis with nominal position $\mathbf{p}^{*}(k-d)$ and a nominal drive speed along the belt $ \dot{\mathbf{p}}^{*}(k-d)$.
Based on the camera characteristics, we sample the delay once per episode as $\Delta t\sim\mathcal{U}(70,150)\,\mathrm{ms}$ and compute $d$ using~\eqref{eq:d}, where $\mathcal{U}(a,b)$ denotes the continuous uniform distribution over $[a,b]$.
Because the conveyor belt is aligned with the y-axis in Cartesian space, we perturb only the $y$-component of the target velocity while keeping $x$ and $z$ components noise-free $\dot{\mathbf{p}}^{*}(k) \leftarrow \dot{\mathbf{p}}^{*}(k) + \mathcal{N}(0,0.05)\ \mathrm{mm/s}$.
At the beginning of each episode, we also randomize the target speed by an additive random offset along belt motion direction (y-axis) $\dot{\mathbf{p}}^{*}(0) \leftarrow \dot{\mathbf{p}}^{*}(0) + \begin{bmatrix} 0 & \mathcal{N}(0,1) & 0 \end{bmatrix}^{\!\top}\ \mathrm{mm/s}$.
For the target position $\tilde{\mathbf{p}}^{*}(k-d)$, we add independent Gaussian noise per axis $\tilde{\mathbf{p}}^{*}(k-d) \leftarrow \tilde{\mathbf{p}}^{*}(k-d) + \mathcal{N}(0,0.5)\ \mathrm{mm}$.
The conveyor belt is driven by a stepper motor with a fixed speed set to $34.9\ \mathrm{mm/s}$; however, the measured speed has noise with a standard deviation of $0.05\ \mathrm{mm/s}$.
Algorithm \ref{alg:residual_sac} summarizes our control method.

\begin{table}[t]
\centering
\begingroup
\caption{SAC training configuration.}
\label{tab:sac_parameters}
\footnotesize
\setlength{\tabcolsep}{3pt}
\renewcommand{\arraystretch}{1.0}
\begin{tabular}{@{}ll@{}}
\hline
Parameter & Value \\
\hline
Hidden layers (actor/each critic) & $2\times256$, ReLU \\
Adam rates (actor/critics; $\alpha$) & $10^{-3}$; $10^{-4}$ \\
Discount $\gamma$ & $0.99$ \\
Initial $\alpha$; target entropy & $0.1$; $-6$ \\
Minibatch; replay capacity & $136$; $1.36\times10^6$ \\
Warm-up / learning starts & $13\,600$ interactions \\
Update schedule after warm-up & $136$ updates / $136$ interactions \\
Training budget & $4000$ tracking episodes \\
\hline
\end{tabular}
\endgroup
\end{table}

Architecture, discount, target mixing and entropy target follow~\cite{SAC2019}; other settings in Table~\ref{tab:sac_parameters} were selected manually using digital-twin tracking performance.
For a new system, test a few candidate hyperparameter values in its digital twin, repeating training with different random seeds.
Select settings that achieve low tracking error without exceeding actuator limits in simulated scenarios not used for training.

\begin{algorithm}[!h]
\caption{Residual zero–shot sim-to-real SAC}
\label{alg:residual_sac}
\begin{algorithmic}[1]
\State \textbf{digital twin inputs:} $P$, $\hat{J}^{\dagger}$, $K_{\text{P}}$, $K_{\text{I}}$, $\delta t$, $\dot{\mathbf{q}}_{\max}$, $\rho$, $L$,  $l$
\State \textbf{offline SAC training in digital twin:} output $\phi^{\ast}$
\vspace{1pt}
\Statex \centerline{\textbf{domain randomization per episode:}}
\vspace{-11pt}
\Statex \hspace{0.6em}(a) kinematic model: additive noise to link positions $\mathcal{N}(0, 0.5\ \mathrm{mm})$, and link orientation $\mathcal{N}(0, 3^\circ)$
\Statex \hspace{0.6em}(b) target speed: $\dot{\mathbf{p}}^{*}(0){\leftarrow}\dot{\mathbf{p}}^{*}(0){+}[0,\mathcal{N}(0,1),0]^\top\ \mathrm{mm/s}$
\Statex \hspace{0.6em}(c) initial joints position: $\mathbf q(0){\leftarrow}\mathbf q(0){+}\mathcal N(0,0.01)\ \mathrm{rad}$
\Statex \hspace{0.6em}(d) initial joints speed: $\dot{\mathbf{q}}(0){\leftarrow}\dot{\mathbf{q}}(0){+}\mathcal N(0,0.001)\ \mathrm{rad/s}$
\Statex \hspace{0.6em}(e) camera delay: $\Delta t\sim\mathcal{U}(70,150)\ \mathrm{ms}$
\Statex \centerline{\textbf{domain randomization per timestep:}}
\vspace{-11pt}
\Statex \hspace{0.6em}(a) joints speed noise: $\dot{\mathbf q}(k){\leftarrow}\dot{\mathbf q}(k){+}\mathcal N(0,0.04)\ \mathrm{rad/s}$
\Statex \hspace{0.6em}(b) target position noise: $\tilde{\mathbf{p}}^{*}(k){\leftarrow}\tilde{\mathbf{p}}^{*}(k){+}\mathcal N(0,0.5)\ \mathrm{mm}$
\Statex \hspace{0.6em}(c) target speed noise:$\dot{\mathbf{p}}^{*}(k){\leftarrow}\dot{\mathbf{p}}^{*}(k){+}\mathcal{N}(0,0.05)\ \mathrm{mm/s}$

\vspace{2pt}
\State \textbf{zero-shot deployment on real robot:} set $k\leftarrow 0$,
\While{in tracking phase}
    \State get target speed from conveyor drive: $\dot{\mathbf{p}}(k)$
    \State get delayed target position from camera: $\mathbf{p}^{*}(k-d)$
    \State get target position from motion planner: $\tilde{\mathbf{p}}^{*}(k)$
    \State update tracking error: $\mathbf e(k){=}\tilde{\mathbf{p}}^{*}(k){-}\mathbf{p}(k)$
    \State get inverse--Jacobian PI controller command:
    \Statex \qquad $\mathbf u(k){=}\hat{J}^{\dagger}(\mathbf q(k))\cdot\left[ \dot{\mathbf{p}}^{*}(k){+}K_{\mathrm P}\mathbf e(k){+}K_{\mathrm I}\!\sum_{i\le k}\mathbf e(i)\delta t\right]$
    \State update $s(k)=\bigl[\mathbf{e}(k),\mathbf{q}(k),\dot{\mathbf{q}}(k),\mathbf{u}(k),\dot{\mathbf{q}}_{\mathrm{SAC}}(k{-}1)\bigr]$
    \State get SAC residual command: $\dot{\mathbf{q}}_{\mathrm{SAC}}(k) = \pi_{\phi^{*}}(s(k))$
    \State apply our residual SAC--PI controller command:
    \Statex \qquad $\dot{\mathbf q}_{\mathrm c}(k){=}\mathbf u(k){+}\dot{\mathbf q}_{\mathrm{SAC}}(k)$, $k\leftarrow k+1$
\EndWhile
\end{algorithmic}
\end{algorithm}

\section{Frequency-Domain Lower Bound on Tracking-Error}\label{sec:IV_lower_bound}
We derive a lower bound on the spectral content of the tracking error for a class of inverse-Jacobian PI controllers.
We define a sliding time window in which we approximate the closed-loop system as a linear time-invariant (LTI) system.
Each window comprises $T_w$ samples, corresponding to a duration $T_{\mathrm{win}} := T_w,\delta t$, and is centered at time index~$k$:
\begin{equation}\label{eq:W_k}
\mathcal{W}_k := {,k{-}\tfrac{T_w}{2},\dots,k{+}\tfrac{T_w}{2},}.
\end{equation}
The window length $T_w=1 \mathrm{s}$ is chosen such that the Jacobian varies slowly,
\begin{equation*}
\frac{|J(\mathbf{q}(k{+}T_w)) - J(\mathbf{q}(k))|_F}
{|J(\mathbf{q}(k))|_F}
\le 5
\end{equation*}
where $|\cdot|_F$ denotes the Frobenius norm.
This allows us to approximate the closed-loop system within $\mathcal{W}_k$ with an LTI system.

We assume the ideal joint model $G_q(e^{j\omega})=\mathbb{I}_6 \in \mathbb{R}^{6\times 6}$ leading to the plant model $P_{\mathbf{q}}(e^{j\omega})=G(e^{j\omega})J(\mathbf{q}(k))$ such that $L_0(e^{j\omega})\approx G(e^{j\omega})C(e^{j\omega})$.
We define the nominal and perturbed sensitivity functions as
\begin{align}
S_{0}(e^{j\omega}) :&= (I + L_{0}(e^{j\omega}))^{-1}, \label{eq:S0}\\
S_{\mathbf{q}}(e^{j\omega}) :&= (I + L_{0}(e^{j\omega}) + \Delta_{L,\mathbf{q}}(e^{j\omega}))^{-1} \notag \\ &= S_{0}(e^{j\omega})(I + \Delta_{L,\mathbf{q}}(e^{j\omega}) S_{0}(e^{j\omega}))^{-1}, \label{eq:S}
\end{align}
where $\Delta_{L,\mathbf{q}}(e^{j\omega}) := P_{\mathbf{q}}(e^{j\omega})\,\Delta_{a}(\mathbf{q}(k))\,C(e^{j\omega})$.
For the sampling period $\delta t$, the frequency range is
$\Omega\subseteq[0,\Omega_{\mathrm N}]$, where
$\Omega_{\mathrm N}=\pi/\delta t\;\mathrm{rad/s}$ is the Nyquist frequency.
We define the admissible inverse-Jacobian PI class 
\begin{equation*}
\begin{aligned}
\mathcal K_{\mathrm{PI}}
:= \bigl\{&K=(K_{\mathrm P},K_{\mathrm I}) \mid{}\;
K_X=\operatorname{diag}(k_{X,1},k_{X,2},k_{X,3}),\\
&0\le k_{X,i}\le 2k_{X,i}^{(0)},\ 
X\in\{\mathrm P,\mathrm I\},\ i=1,2,3
\bigr\},
\end{aligned}
\end{equation*}
such that $k_{X,i}^{(0)}$ denotes the $i$th loop-shaped baseline gain from
Section~\ref{subsec:gain_optim}.
Using \eqref{eq:S} and the singular-value and matrix-norm inequalities~\cite{horn2012matrix} yields
\begin{equation}\label{eq:sigma_min_S}
\begin{split}
\sigma_{\min}\!\bigl(S_{\mathbf{q}}(e^{j\omega})\bigr)\;&\ge\; \gamma_{\mathbf{q}}(\omega;K),\\
\gamma_{\mathbf{q}}(\omega;K)
  &:=
\frac{\sigma_{\min}(S_0(e^{j\omega}))}
     {1+\|\Delta_{L,\mathbf{q}}(e^{j\omega}) S_0(e^{j\omega})\|_2 },
\end{split}
\end{equation}
where $K\in\mathcal K_{\mathrm{PI}},\ \omega\in\Omega$, $\sigma_{\min}(\cdot)$ denotes the smallest singular value, and $\|\cdot\|_2$ is the induced spectral $2$-norm.

Within each window $\mathcal{W}_k$, we assume that the feedforward velocity $\dot{\mathbf{p}}^{*}(k)$ is deterministic;
we then decompose the reference position and tracking error \begin{align*}
\tilde{\mathbf p}^{*}(k')
   =& \boldsymbol\mu_k(k') + \mathbf{v}(k'),\\
   \mathbf e(k') =& \mathbf e_{\mathrm{det}}(k') + \mathbf e_{\mathrm{st}}(k'),
\end{align*}
where $\forall\, k' \in \mathcal{W}_k$.
$\boldsymbol\mu_k(k')$ and $\mathbf{v}(k')$ are the locally deterministic and the zero-mean wide-sense stationary (WSS) components of the reference position signal, respectively.
$\mathbf e_{\mathrm{det}}(k')$ denotes the deterministic response to $\boldsymbol\mu_k(k')$ and $\dot{\mathbf{p}}^{*}(k)$, and $\mathbf e_{\mathrm{st}}(k')$ is the stochastic error obtained by driving the LTI system with $\mathbf v(k')$.

We define the power spectral density (PSD) of the stochastic reference and tracking-error components \cite{oppenheim1999discrete} as
\begin{equation*}
\Phi_v(\omega)
:= \frac{\delta t}{2\pi}\mathbb{E}\!\bigl[\|\mathbf V(\omega)\|_2^2\bigr],
\ 
\Phi_{e,\mathrm{st}}(k,\omega;K)
:= \frac{\delta t}{2\pi}\mathbb{E}\!\bigl[\|\mathbf E_{\mathrm{st}}(\omega)\|_2^2\bigr],
\end{equation*}
where $\mathbf V(\omega)$ and $\mathbf E_{\mathrm{st}}(\omega)$ are the
Fourier transforms of $\mathbf v(k')$ and $\mathbf e_{\mathrm{st}}(k')$,
respectively.
For the LTI system,
$\mathbf E_{\mathrm{st}}(\omega)
=S_{\mathbf q}(e^{j\omega})\mathbf V(\omega)$.
Moreover, \eqref{eq:sigma_min_S} implies $S_{\mathbf q}(e^{j\omega})^*
S_{\mathbf q}(e^{j\omega})
\succeq\gamma_{\mathbf q}(\omega;K)^2\mathbb{I}_3,$ where $(\cdot)^*$ denotes the conjugate transpose.
Therefore, taking expectations yields \cite{horn2012matrix} $\mathbb E\!\left[\|\mathbf E_{\mathrm{st}}(\omega)\|_2^2\right]
\ge
\gamma_{\mathbf q}(\omega;K)^2
\mathbb E\!\left[\|\mathbf V(\omega)\|_2^2\right]$.
Consequently, given $K\in\mathcal K_{\mathrm{PI}},\ \omega\in\Omega$ we can write
\begin{equation}\label{eq:pre_taper_LB_PSD}
\Phi_{e,\mathrm{st}}(k,\omega;K)
\;\ge\;
\Phi_{e,\mathrm{st},\mathrm{LB}}(k,\omega;K)
:= \gamma_{\mathbf q}(\omega;K)^2\,\Phi_v(\omega).
\end{equation}

We apply a deterministic taper $w(k')$ within each window $\mathcal W_k$ to reduce spectral leakage caused by the implicit periodic extension of finite data segments \cite{percival1993spectral}.
We use the Hann taper defined as
\begin{equation*}
w(k')
:= 
\frac12\!\left(1 - 
\cos\!\left(\frac{2\pi(k' - k + T_w/2)}{T_w-1}\right)\right),
\quad
k' \in \mathcal W_k.
\end{equation*}
The tapered stochastic error is then $\mathbf e_{\mathrm{st},w}(k') 
:= w(k')\,\mathbf e_{\mathrm{st}}(k')$, $\forall\, k' \in \mathcal W_k$.
The corresponding tapered PSD is
\begin{equation*}
\Phi_{e,\mathrm{st}}^w(k,\omega;K)
:= 
\frac{\delta t}{2\pi}\mathbb E\!\left[
\frac{1}{U}\,
\bigl\|\mathbf E_{\mathrm{st},w}(\omega)\bigr\|_2^2
\right],
\ 
U:=\!\sum_{k' \in \mathcal W_k}\!w(k')^2,
\end{equation*}
where $\mathbf E_{\mathrm{st},w}(\omega)$ is the Fourier transform of 
$\mathbf e_{\mathrm{st},w}(k')$.
For the zero-mean WSS error, the expected tapered PSD is
\cite{percival1993spectral}
\begin{equation*}
\Phi_{e,\mathrm{st}}^w(k,\omega;K)
=
\int_{-\pi}^{\pi}
\Phi_{e,\mathrm{st}}(k,\theta;K)\,
H_w(\omega-\theta)\,\mathrm d\theta,
\end{equation*}
where $H_w(\omega):=\frac{|W(\omega)|^2}{2\pi U}\ge 0$ and $W(\omega)$ is the Fourier transform of $w(k')$.
According to
\eqref{eq:pre_taper_LB_PSD}, we have:
\begin{equation*}
\Phi_{e,\mathrm{st}}^w(k,\omega;K)
\ge
\Phi_{e,\mathrm{st},\mathrm{LB}}^w(k,\omega;K),
\end{equation*}
where $\Phi_{e,\mathrm{st},\mathrm{LB}}^w(k,\omega;K)
:=
\int_{-\pi}^{\pi}
\Phi_{e,\mathrm{st},\mathrm{LB}}(k,\theta;K)\,
H_w(\omega-\theta)\,\mathrm d\theta.$
Since the deterministic component $\mathbf e_{\mathrm{det}}(k')$ contributes nonnegative expected spectral energy, we have
\begin{equation}\label{eq:win_lb_PSD}
\Phi_{e}^w(k,\omega;K)
\;\ge\;
\Phi_{e,\mathrm{st}}^w(k,\omega;K)
\;\ge\;
\Phi_{e,\mathrm{st},\mathrm{LB}}^w(k,\omega;K),
\end{equation}
where $K\in\mathcal K_{\mathrm{PI}},\ \omega\in\Omega$, and $\Phi_{e}^w(k,\omega;K)$ is the expected tapered short-time spectral estimate of the full error.

The pointwise lower bound common to the entire admissible PI class is
\begin{equation}
\Phi_{e,\mathrm{LB}}^{*}(k,\omega)
:=\inf_{K\in\mathcal K_{\mathrm{PI}}}
\Phi_{e,\mathrm{st},\mathrm{LB}}^w(k,\omega;K).\label{eq:optimal_LB_PI_class}
\end{equation}
Accordingly, \eqref{eq:win_lb_PSD} gives
\begin{equation}\label{eq:win_lb_PSD_PIclass}
\Phi_{e}^w(k,\omega;K)
\;\ge\;
\Phi_{e,\mathrm{LB}}^{*}(k,\omega),
\quad K\in\mathcal K_{\mathrm{PI}},\ \omega\in\Omega.
\end{equation}
Under the frozen-LTI and ideal-joint-model assumptions, with deterministic
feedforward velocity and a zero-mean WSS reference component,
$\Phi_{e,\mathrm{LB}}^{*}(k,\omega)$ is a pointwise lower bound on the
average tapered tracking-error power spectrum for every
$K\in\mathcal K_{\mathrm{PI}}$.
In the following subsections, we compute $\Phi_e^w(k,\omega)$ from the measured tracking error of our hybrid controller and compare it directly with $\Phi_{e,\mathrm{LB}}^{*}(k,\omega)$.

\section{Performance Evaluation}\label{sec:V_performance_evaluation}
To evaluate robustness to kinematic mismatch, we use a biased kinematic
model in the model-based controller in both simulation and real-robot
experiments.
We introduce the bias by perturbing each joint frame's position and orientation with random offsets of approximately $\pm0.5\,\mathrm{mm}$ and $\pm3^{\circ}$, respectively.

Next, we report the tracking results in simulation and on the real robot and evaluate our residual SAC--PI controller on hardware at different conveyor speeds.

\subsection{Numerical Results}
\begin{figure}[!h] 
\centering
\includegraphics[width=.95\columnwidth, trim=0.5cm 0.5cm 0.5cm 0.5cm, clip]{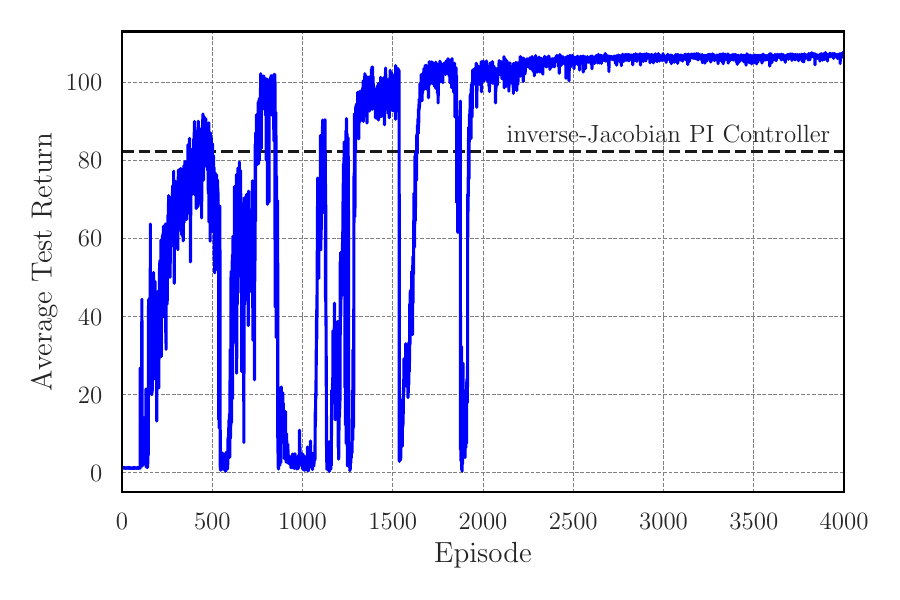}
\caption{Learning curve of the SAC controller.
The maximum return is $110$.
The dashed line shows the inverse–Jacobian PI controller performance.}
\label{fig:learning_curve_SAC_FEP_Pybullet}
\end{figure}

Using PyBullet, we train SAC for $4000$ episodes in the tracking phase, lasting $N = 110$ time instants with a sampling time of $\delta t = 100~\mathrm{ms}$.
Fig. \ref{fig:learning_curve_SAC_FEP_Pybullet} shows the SAC learning curve.
After $2000$ episodes, our residual SAC--PI controller consistently outperforms the inverse–Jacobian PI baseline.
Training uses a nominal target speed of $34.9~\mathrm{mm/s}$ with episode-wise Gaussian speed offsets of standard deviation $1~\mathrm{mm/s}$
(Section~\ref{subsec:residual_rl}).
To assess generalization beyond these small training variations, Section~\ref{subsec:experiment_results} evaluates the same trained policy with unchanged controller settings at different target speeds (Table~\ref{table:real_methods_comparison_diff_speeds}).

\begin{figure}[!h] 
\centering
\includegraphics[width=.95\columnwidth,trim=0.2cm 0.3cm 0.2cm 0.2cm, clip]{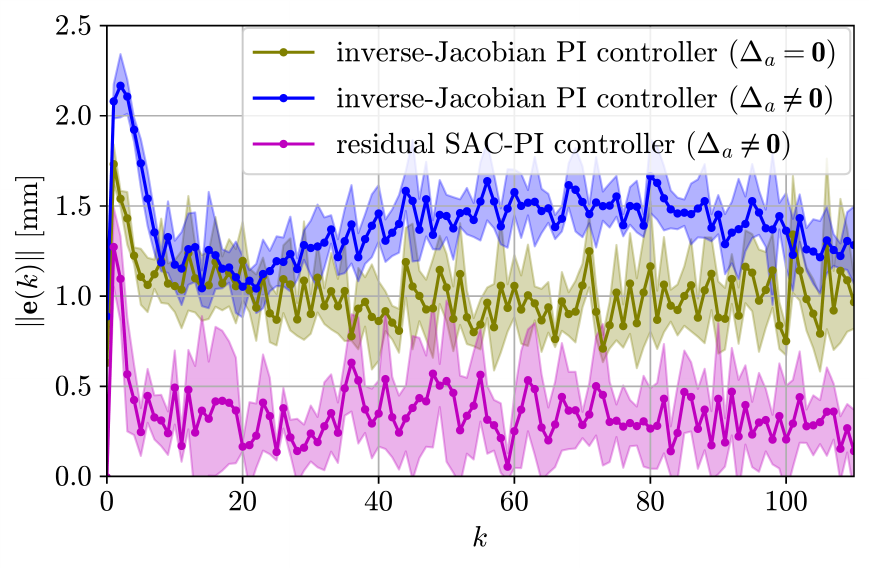}
\caption{Monte Carlo study tracking error mean and $95\%$ confidence interval in the simulation. \label{fig:simulation_tracking_errors_compare_methods}}

\end{figure}

We compare our controller with an inverse-Jacobian PI controller using both biased and true kinematic models.
Given the true kinematic model ($\Delta_a=0$), we reoptimize the inverse–Jacobian with PI controller gains as explained in \ref{subsubsec:controller_gains_design}, resulting in $K_{\mathrm{P}}=\mathrm{diag}(3.7,10.9,8.6)$ and $K_{\mathrm{I}}=\mathrm{diag}(3.0,3.4,7.7)$.
We perform a Monte Carlo study with $M=50$ episodes, each with different initial conditions and domain randomization.
Each episode begins after its startup phase and includes the tracking phase. 
In Fig. \ref{fig:simulation_tracking_errors_compare_methods}, our controller's mean tracking error is around $0.4\, \mathrm{mm}$.
The inverse–Jacobian PI controllers with or without kinematic bias exhibit larger tracking errors, around $1.4\,\mathrm{mm}$ and $1.0\,\mathrm{mm}$, respectively.
Although the true kinematic model improves performance, the corrected inverse–Jacobian PI controller still performs worse than our controller, showing that SAC compensates for kinematic errors, measurement noise, and unmodeled joint-drive dynamics.

The confidence intervals in Fig.~\ref{fig:simulation_tracking_errors_compare_methods} are similar across all controllers.
Domain randomization induces mild variations in the closed-loop dynamics, which accounts for the observed spread.
The SAC residual effectively adapts to these variations and significantly reduces the mean tracking error.
This small uncertainty can be further mitigated through smoother residual filtering or more consistent delay compensation.

\begin{figure}[!h] 
\centering
\includegraphics[width=.95\columnwidth,trim=0.2cm 0.3cm 0.3cm 0.2cm, clip]{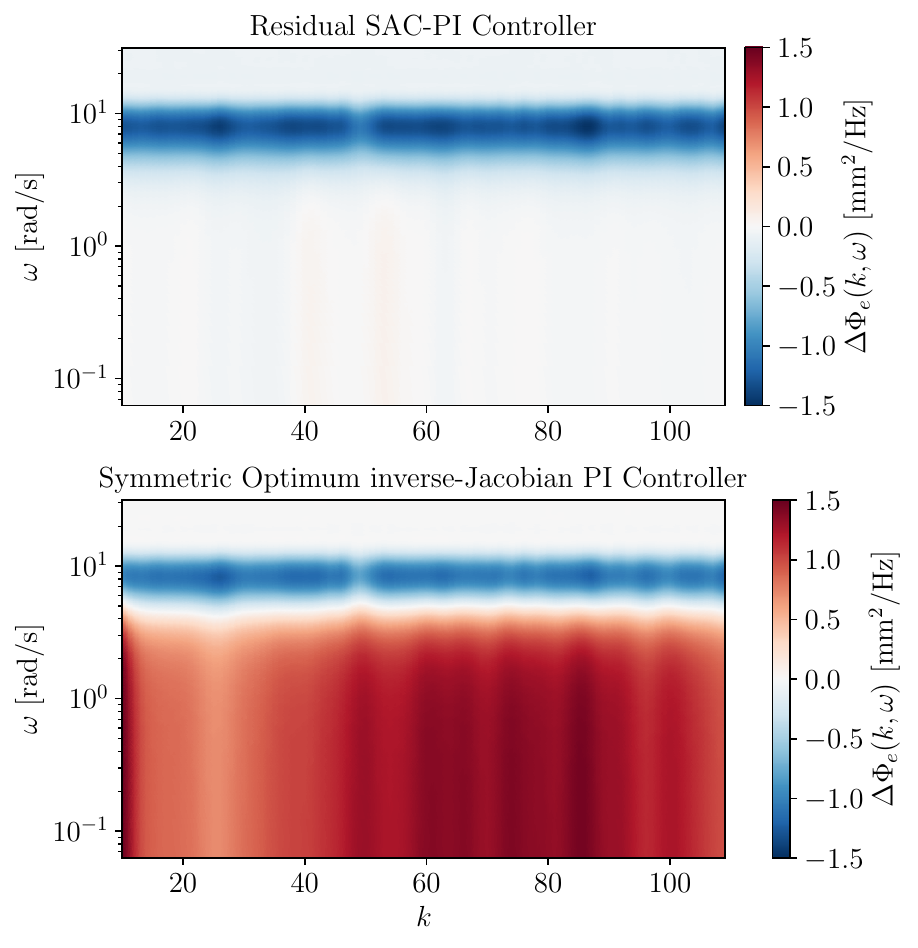}
\caption{Simulated error PSD deviation from the optimal PI-class lower bound for the inverse–Jacobian PI class of controllers \label{fig:PSD_compare_LB}}

\end{figure}

Given measured tracking-error sequence $\mathbf{e}(k)$ for $k\in[0,\dots,N]$, we compute its tapered PSD $\Phi_{e}^{w}(k,\omega)$ over each sliding window~$\mathcal{W}_k$. 
The deviation from the optimal PI-class lower bound in \eqref{eq:optimal_LB_PI_class} is defined as
\begin{equation}\label{eq:delta_phi_c}
\Delta\Phi_{e}(k,\omega):= \Phi_{e}^{w}(k,\omega)-\Phi_{e,\mathrm{LB}}^\star(k,\omega),
\quad \forall\omega\in\Omega,
\end{equation}
quantifying, within each window~$\mathcal{W}_k$, the excess spectral error power above the windowed tracking performance lower bound for controllers in $\mathcal{K}_{\mathrm{PI}}$.  
Fig. \ref{fig:PSD_compare_LB} shows that our residual SAC--PI controller exhibits negative PSD deviations at low frequencies, with a consistently negative accumulation of PSD deviation across all frequencies at all time instants, i.e., $\sum_{\omega \in \Omega}\Delta\Phi_{e}(k,\omega)<0,\, \forall k\in\mathcal{W}$
This suggests a reduced tracking error relative to the lower bound for PI-class controllers. 
In contrast, the inverse-Jacobian PI controller with symmetric optimum gains using the same biased kinematics model shows positive PSD deviations at low frequencies and positive cumulative values across several time instants, as expected from the lower bound.

\subsection{Experimental Results}\label{subsec:experiment_results}
\begin{figure}[!h] 
\centering
\includegraphics[width=.95\columnwidth, trim=0.2cm 0.3cm 0.2cm 0.2cm, clip]{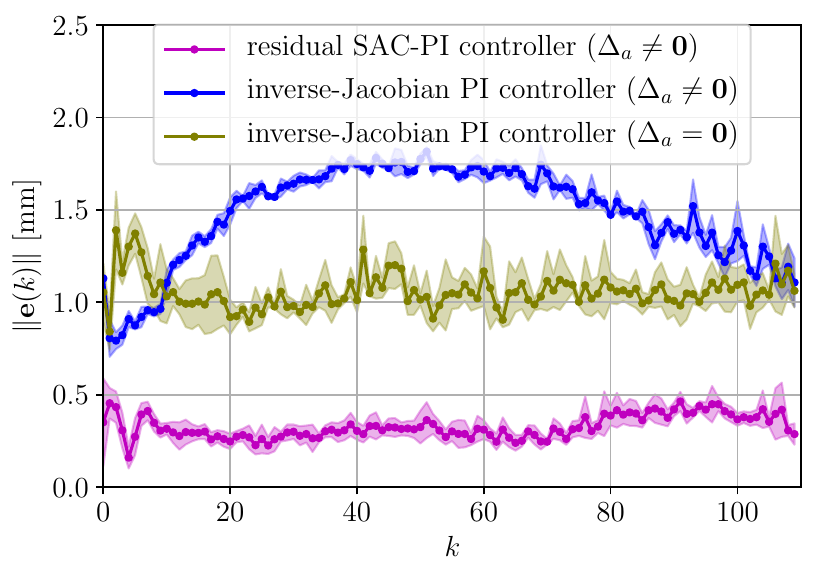}
\caption{Tracking error comparison on a real robot system between inverse–Jacobian PI controllers and our residual SAC--PI controller.
Shaded area shows the $95\%$ confidence interval.}
\label{fig:real_tracking_errors_compare}



\end{figure}

We evaluate our residual SAC--PI controller on the real Franka Emika Panda arm, where we train the SAC policy entirely in simulation without any real-system data.
The inverse–Jacobian PI baseline uses a biased kinematic model with position and orientation errors of about $0.5\,\mathrm{mm}$ and $3^\circ$, respectively.
We also consider another inverse–Jacobian PI controller which uses the true kinematic model ($\Delta_a=0$).
We use the already designed symmetric PI gains for both inverse-Jacobian PI controllers.
We compare the tracking performance of both inverse-Jacobian PI controllers against our residual SAC--PI controller, which also uses the biased kinematic model.
We repeat our tracking experiments $M=100$ times.
Fig. \ref{fig:real_tracking_errors_compare} shows that our method keeps average tracking errors under $0.5\ \mathrm{mm}$, while the inverse–Jacobian PI controller with a biased pseudo-inverse Jacobian exceeds $1.5\ \mathrm{mm}$.
The symmetric optimum inverse-Jacobian PI controller with the true kinematic model also performs worse than our residual SAC--PI controller, with a $1.0\ \mathrm{mm}$ error.
This shows that on the real system, our controller can overcome not only kinematic model inaccuracy but also measurement noise and dynamics model mismatches, further reducing the tracking error.

\begin{figure}[!h] 
\centering
\includegraphics[width=.95\columnwidth,trim=0.2cm 0.3cm 0.3cm 0.2cm, clip]{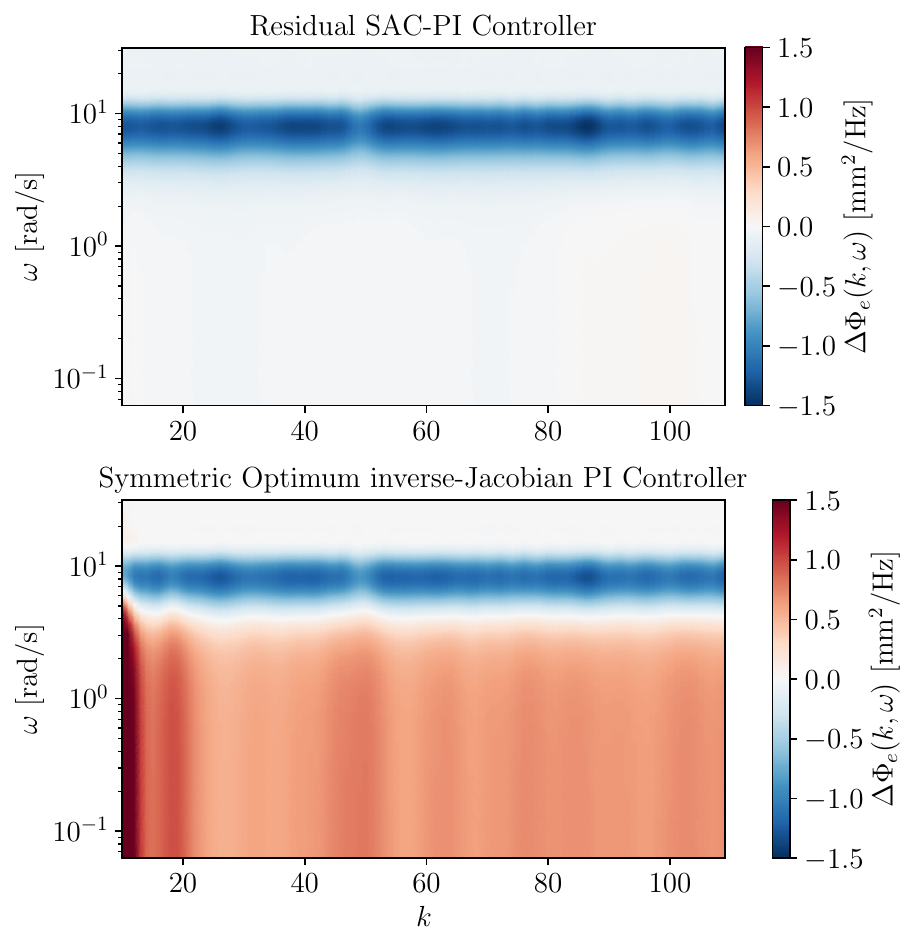}
\caption{Measured error PSD deviation from the optimal PI-class lower bound in \eqref{eq:delta_phi_c} for the inverse–Jacobian PI controller and our residual SAC--PI controller \label{fig:PSD_compare_LB_real}}

\end{figure}

In Fig. \ref{fig:PSD_compare_LB_real}, we compare the tracking performance of the residual SAC--PI controller and the baseline inverse–Jacobian PI controller, both using the biased kinematic model, against the lower bound in \eqref{eq:win_lb_PSD}.
Our controller shows reduced spectral power relative to the bound at low frequencies, with consistently negative cumulative deviations across all frequencies at every time instant, reflecting superior tracking performance. 
In contrast, the inverse–Jacobian PI controller shows positive low-frequency deviations, indicating it cannot achieve tracking errors below the lower bound under kinematic and measurement uncertainties.

To evaluate generalization across target speeds, Table~\ref{table:real_methods_comparison_diff_speeds} compares the residual SAC--PI controller with the biased-kinematics inverse--Jacobian PI controller at three fixed conveyor speeds using symmetric optimum gains.
We use the same trained SAC policy and controller settings at all speeds, without retraining, fine-tuning, or speed-specific tuning.
We report the \emph{mean L2 tracking error} 
\begin{equation}\label{eq:mean_e_L2}
    \varepsilon := \frac{1}{M \cdot N} \sum_{m=1}^{M}\sum_{k=1}^{N} \|\mathbf{e}_m(k)\|,
\end{equation}
where $\|\mathbf{e}_m(k)\|$ is Euclidean norm of tracking error at time index $k$ and episode $m$.
Given the standard deviation of the tracking error at time instant $k$ over $M$ episodes as $\eta(k)$, we denote the tracking error uncertainty by
\begin{equation}
\eta := \frac{1}{N} \sum_{k=1}^{N} \eta(k),
\end{equation}
which quantifies the uncertainty across episodes.
Additionally, we compare the root mean square (RMS) of the error on the entire reference tracking trajectory defined as 
\begin{equation}\label{eq:e_RMS}
e_{\text{RMS}} := \sqrt{ \frac{1}{N \cdot M} \sum_{m=1}^{M} \sum_{k=1}^{N} \|\mathbf{e}_m(k)\|^2 }.
\end{equation}
Together, these metrics provide a comprehensive evaluation of tracking accuracy and robustness across methods and speeds.

\setlength{\tabcolsep}{5pt}
\begin{table}[thb]
\centering
\setlength\extrarowheight{1.1pt}
\caption {\label{table:real_methods_comparison_diff_speeds} Tracking error metrics of different controllers given different conveyor belt speeds $\dot{\mathbf{p}}^{\star}$} 
\begin{tabular*}{0.99\columnwidth}{@{\extracolsep{\fill}}ccccc} \toprule
  {} & {$\dot{\mathbf{p}}^{\star}$}  & {$\varepsilon$} & {$\eta$} & {$e_{\text{RMS}}$}  \\ 
  Controller          &   $\mathrm{mm/s}$   & $\mathrm{mm}$ & $\mathrm{mm}$ & $\mathrm{mm}$ \\ \toprule
inverse–Jacobian PI          & 27.9     & 1.1 & 0.1 & 1.2 \\
residual SAC--PI          & 27.9      & 0.7 & 0.1 & 0.7  \\
\midrule[0.1pt]
 inverse–Jacobian PI              & 34.9      & 1.2  & 0.1 & 1.2\\
 residual SAC--PI      & 34.9      &  0.3 & 0.1 & 0.4\\
 \midrule[0.1pt]
 inverse–Jacobian PI       & 41.9      & 1.4 & 0.1 & 1.5 \\
 residual SAC--PI       & 41.9      & 0.4 & 0.1 & 0.5 \\
 \bottomrule  
\end{tabular*}
\end{table}

Across all speeds, the proposed residual SAC--PI controller consistently outperformed the baseline inverse–Jacobian PI controller. 
While the baseline controller's mean L2 tracking error increased with belt speed from $1.1\ \mathrm{mm}$ to $1.4\ \mathrm{mm}$, the residual SAC--PI controller maintained sub-millimeter accuracy, with mean errors below $0.7\ \mathrm{mm}$. 
The RMS values confirm this improvement, showing a reduction of about $0.8\ \mathrm{mm}$ ($e_{\mathrm{RMS}}=0.4$–$0.7\ \mathrm{mm}$ versus $1.2$–$1.5\ \mathrm{mm}$). 
The comparable uncertainty $\eta$ between controllers indicates that the improvement stems from systematic error reduction rather than increased variability.

\section{Conclusion}\label{sec:VI_Conclusion}
We show how reinforcement learning can enhance a model-based controller despite low-fidelity simulation and sensing uncertainties. 
We trained the SAC policy entirely in simulation, with domain randomization over the kinematic model, delay, noise, and target motion.
Under a biased kinematic model and within each local time window, we derive a short-time lower bound on the tracking error spectrum across frequencies for a class of model-based controllers.
Real-robot experiments confirmed that our hybrid controller achieves improved tracking performance below the lower bound. 
Future work will analyze the directional effects of modeling mismatch across task-space axes and develop adaptive mismatch learning for improved sim-to-real compensation.

\bibliographystyle{IEEEtran}
\bibliography{main}

\end{document}